# Max-Margin Nonparametric Latent Feature Models for Link Prediction

**Jun Zhu**                                                                 DCSZJ@MAIL.TSINGHUA.EDU.CN

State Key Lab of Intelligent Tech & Sys, Tsinghua National TNList Lab,
Department of Computer Science and Technology, Tsinghua University, Beijing, 100084 China

## Abstract

We present a max-margin nonparametric latent feature relational model, which unites the ideas of max-margin learning and Bayesian nonparametrics to discover discriminative latent features for link prediction and automatically infer the unknown latent social dimension. By minimizing a hinge-loss using the linear expectation operator, we can perform posterior inference efficiently without dealing with a highly nonlinear link likelihood function; by using a fully-Bayesian formulation, we can avoid tuning regularization constants. Experimental results on real datasets appear to demonstrate the benefits inherited from max-margin learning and fully-Bayesian nonparametric inference.

## 1. Introduction

As the availability and scope of social networks and relational datasets increase, a considerable amount of attention has been devoted to the statistical analysis of such data, which is typically represented as a graph in which the vertices represent entities and edges represent links between entities. Link prediction is one fundamental problem in analyzing these social network or relational data, and its goal is to predict unseen links between entities given the observed links. Often there is extra information about links and entities such as attributes and timestamps (Liben-Nowell & Kleinberg, 2003; Backstrom & Leskovec, 2011; Miller et al., 2009) that can be used to help with prediction.

Recently, various approaches based on probabilistic models have been developed for link prediction. One class of such models utilize a latent feature matrix and a link function (e.g., the commonly used sigmoid function) (Hoff, 2007; Miller et al., 2009) to define the link formation probability distribution. These latent feature models were shown to generalize latent class (Nowicki & Snijders, 2001; Airoldi et al., 2008) and latent distance (Hoff et al., 2002) models and are thus able to represent both homophily and stochastic equivalence, which are important properties commonly observed in real-world social network and relational data. The parameters for these probabilistic models are typically estimated with MLE or their posterior distributions are inferred with Monte Carlo methods. Such techniques have demonstrated competitive results on various datasets. However, to determine the unknown dimensionality of the latent feature space (or latent social space), most of the existing approaches rely on a general model selection procedure, e.g., cross-validation, which could be expensive by comparing many different settings. The work (Miller et al., 2009) is an exception, which presents a nonparametric Bayesian method to automatically infer the unknown social dimension.

This paper presents an alternative way to develop nonparametric latent feature relational models. Instead of defining a normalized link likelihood model, we propose to directly minimize some objective function (e.g., hinge-loss) that measures the quality of link prediction, under the principle of maximum entropy discrimination (MED) (Jaakkola et al., 1999; Jebara, 2002), which was introduced as an elegant framework to integrate max-margin learning and Bayesian generative modeling. The present work extends MED in several novel ways to solve the challenging link prediction problem. First, like (Miller et al., 2009), we use nonparametric Bayesian techniques to automatically resolve the unknown dimension of a latent social space, and thus our work represents an attempt towards uniting Bayesian nonparametrics and max-margin learning, which have been largely treated as two isolated topics. Second, we present a fully-Bayesian method to avoid tuning regularization constants. By minimizing a hinge-loss, our model avoids dealing with a highly nonlinear link likelihood (e.g., sigmoid) and can be efficiently solved using variational methods, where the sub-problems of max-margin learning are solved with existing high-performance solvers. Experimental results on three real datasets appear to demonstrate that 1) using max-margin learning can significantly improve the link prediction performance, and 2) using fully-Bayesian methods, we can avoid tuning regularization constants without sacrificing the performance, and dramatically decrease running time.

The paper is structured as follows. Sec 2 introduces existing latent feature models, as well as a new insight about the connections between these models. Sec 3 presents the max-margin latent feature relational model and a fully-Bayesian formulation. Sec 4 presents empirical results. Finally, Sec 5 concludes.

## 2. Latent Feature Relational Models

Assume we have an $N \times N$ relational link matrix $Y$, where $N$ is the number of entities. We consider the binary case, where the entry $Y_{ij} = +1$ (or $Y_{ij} = -1$) indicates the presence (or absence) of a link between entity $i$ and entity $j$. We emphasize that all the latent feature models introduced below can be extended to deal with real or categorical $Y$. $Y$ is not fully observed. The goal of link prediction is to learn a model from observed links such that we can predict the values of unobserved entries of $Y$. In some cases, we may have observed attributes $X_{ij} \in \mathbb{R}^D$ that affect the link between $i$ and $j$.

In a latent feature relational model, each entity is associated with a vector $\mu_i \in \mathbb{R}^K$, a point in a latent feature space (or latent social space). Then, the link likelihood is generally defined as

$$p(Y_{ij} = 1 | X_{ij}, \mu_i, \mu_j) = \Phi(\mu + \boldsymbol{\beta}^\top X_{ij} + \psi(\mu_i, \mu_j)), \quad (1)$$

where a common choice of $\Phi$ is the sigmoid function, i.e., $\Phi(t) = \frac{1}{1+e^{-t}}$. For the latent distance model (Hoff et al., 2002), we have

$\psi(\mu_i, \mu_j) = -d(\mu_i, \mu_j)$, where $d$ is a distance function.

For the latent eigenmodel (Hoff, 2007), which generalizes the latent distance model and the latent class model for modeling symmetric relational data, we have

$\psi(\mu_i, \mu_j) = \mu_i^\top D \mu_j$, where $D$ is a diagonal matrix.

In the above models, the dimension $K$ is unknown a *priori*, and a model selection procedure (e.g., cross-validation) is needed. The nonparametric latent feature relational model (LFRM) (Miller et al., 2009) leverages the recent advances in Bayesian nonparametrics to automatically infer the latent dimension. Moreover, LFRM differs from the above models by inferring binary latent features and defining

$$\psi(\mu_i, \mu_j) = \mu_i^\top W \mu_j, \text{ where } \mu_i \in \{0, 1\}^\infty.$$

We will use $Z$ to denote a binary feature matrix, where each row corresponds to the latent feature of an entity. For LFRM, we have $Z = [\mu_1^\top; \cdots; \mu_N^\top]$. Fully-Bayesian inference with MCMC sampling is usually performed for these models by imposing appropriate priors on latent features and model parameters. In LFRM, Indian buffet process (IBP) (Griffiths & Ghahramani, 2006) was used as the prior of $Z$ to induce a sparse latent feature vector for each entity.

Miller et al. discussed the expressiveness of LFRM over latent class models. Here, we provide another support for the expressiveness. For modeling symmetric relational data, we usually constrain $W$ to be symmetric (Miller et al., 2009). Since a symmetric real matrix is diagonalizable, i.e., there exists an orthogonal matrix $Q$ satisfying that $Q^\top W Q$ is a diagonal matrix, denoted again by $D$, we have $W = QDQ^\top$. Thus we can treat $ZQ$ as the effective real-valued latent features and conclude that *LFRM subsumes the latent eigenmodel for modeling symmetric relational data*.

## 3. Max-margin Latent Feature Models

Now, we present the max-margin latent feature model for link prediction. We first briefly review the basic concepts of MED (Jaakkola et al., 1999; Jebara, 2002).

### 3.1. MED

We consider binary classification, where the response variable $Y$ takes values from $\{+1, -1\}$. Let $X$ be an input feature vector and $F(X; \eta)$ be a discriminant function parameterized by $\eta$. Let $\mathcal{D} = \{(X_n, Y_n)\}_{n=1}^N$ be a training set and define $h_\ell(x) = \max(0, \ell - x)$, where $\ell$ is a positive cost parameter. Unlike standard SVMs, which estimate a single $\eta$, MED learns a distribution $p(\eta)$ by solving an entropic regularized risk minimization problem with prior $p_0(\eta)$

$$\min_{p(\eta)} \ \mathrm{KL}(p(\eta) \| p_0(\eta)) + C \mathcal{R}(p(\eta)), \quad (2)$$

where $C$ is a positive constant; $\mathrm{KL}(p\|q)$ is the KL divergence; $\mathcal{R}(p(\eta)) = \sum_n h_1(Y_n \mathbb{E}_{p(\eta)}[F(X_n; \eta)])$ is the hinge-loss that captures the large-margin principle underlying the MED prediction rule

$$\hat{Y} = \mathrm{sign} \mathbb{E}_{p(\eta)}[F(X; \eta)]. \quad (3)$$

By defining $F$ as the log-likelihood ratio of a Bayesian generative model, MED provides an elegant way to integrate the discriminative max-margin learning and Bayesian generative modeling. MED subsumes SVM as a special case and has been extended to incorporate latent variables (Jebara, 2002; Zhu et al., 2009) and to perform structured output prediction (Zhu & Xing, 2009). Recent work has further extended MED to unite Bayesian nonparametrics and max-margin learning (Zhu et al., 2011a;b), which have been largely treated as isolated topics, for learning better classification models. The present work contributes by introducing a novel generalization of MED to perform the challenging relational link prediction.

### 3.2. MED Latent Feature Relational Model

Now, we present the max-margin latent feature model for link prediction. Based on the above discussions, we use the same formulations as the most general LFRM model. Specifically, we represent each entity using a set of binary features and let $Z$ to denote the binary feature matrix, of which each row corresponds to an entity and each column corresponds to a feature. The entry $Z_{ik} = 1$ means that entity $i$ has the feature $k$.

If the features $Z_i$ and $Z_j$ are given, we can naturally define the *latent discriminant function* as

$$f(Z_i, Z_j; X_{ij}, W, \eta) = Z_i W Z_j^\top + \eta^\top X_{ij} \quad (4)$$
$$= \mathrm{Tr}(W Z_j^\top Z_i) + \eta^\top X_{ij},$$

where $W$ is a real-valued matrix and the entry $W_{kk'}$ is the weight that affects the link from entity $i$ to entity $j$ if entity $i$ has feature $k$ and entity $j$ has feature $k'$. For finite sized matrices $Z$ with $K$ columns, we can define the prior as a Beta-Bernoulli process (Meeds et al., 2007). In the infinite case, where $Z$ has an infinite number of columns, we adopt the Indian buffet process (IBP) prior over the unbounded binary matrices as described in (Griffiths & Ghahramani, 2006).

Let $\Theta = \{W, \eta\}$ be all the parameters. To make this model Bayesian, we also treat $\Theta$ as random, with a prior $p_0(\Theta)$. To make prediction, we need to get rid of the uncertainty of latent variables, and we define the *effective discriminant function* as an expectation

$$f(X_{ij}) = \mathbb{E}_{p(Z,\Theta)}[f(Z_i, Z_j; X_{ij}, \Theta)]. \quad (5)$$

Then, the prediction rule is $\hat{Y}_{ij} = \mathrm{sign} f(X_{ij})$. Let $\mathcal{I}$ be the set of pairs that have observed links. The hinge loss of the expected prediction rule is

$$\mathcal{R}_\ell(p(Z, \Theta)) = \sum_{(i,j) \in \mathcal{I}} h_\ell(Y_{ij} f(X_{ij})), \quad (6)$$

Let $p_0(Z)$ be the prior on the latent feature matrix. We define the MED latent feature relational model (MedLFRM) as solving the problem

$$\min_{p(Z,\Theta) \in \mathcal{P}} \mathrm{KL}(p(Z,\Theta) \| p_0(Z,\Theta)) + C \mathcal{R}_\ell(p(Z,\Theta)) \quad (7)$$

In graphical models, it is well known that introducing auxiliary variables could simplify the inference by converting marginal dependence into conditional independence. Here, we follow this principle and introduce additional variables for the IBP prior $p_0(Z)$. One elegant way to do that is the stick-breaking representation of IBP (Teh et al., 2007). Specifically, let $\pi_k \in (0,1)$ be a parameter associated with column $k$ of $Z$. The parameters $\boldsymbol{\pi}$ are generated by a stick-breaking process, that is, $\pi_1 = \nu_1$, and $\pi_k = \nu_k \pi_{k-1} = \prod_{i=1}^k \nu_i$, where $\nu_i \sim \mathrm{Beta}(\alpha, 1)$. Given $\pi_k$, each $Z_{nk}$ in column $k$ is sampled independently from Bernoulli($\pi_k$). This process results in a decreasing sequence of probabilities $\pi_k$, and the probability of seeing feature $k$ decreases exponentially with $k$ on a finite dataset. With this representation, we have the augmented MedLFRM

$$\min_{p(\boldsymbol{\nu},Z,\Theta)} \mathrm{KL}(p(\boldsymbol{\nu},Z,\Theta) \| p_0(\boldsymbol{\nu},Z,\Theta)) + C\mathcal{R}_\ell(p(Z,\Theta)) \quad (8)$$

where $p_0(\boldsymbol{\nu}, Z, \Theta) = p_0(\boldsymbol{\nu}) p(Z|\boldsymbol{\nu}) p_0(\Theta)$.

We make two comments about the above definitions. First, we have adopted the similar method as in (Zhu et al., 2011a;b) to define the discriminant function using the expectation operator, instead of the traditional log-likelihood ratio of a Bayesian generative model with latent variables (Jebara, 2002; Lewis et al., 2006). The linearity of expectation makes our formulation simpler than the one that could be achieved by using a highly nonlinear log-likelihood ratio. Second, although a likelihood model can be defined as in (Zhu et al., 2011a;b) to perform hybrid learning, we have avoided doing that because the sigmoid link likelihood model in Eq. (1) is highly nonlinear and it could make the hybrid problem hard to solve.

#### 3.2.1. Inference with Truncated Mean-Field

The above problem has nice properties. For example, $\mathcal{R}_\ell$ is a piece-wise linear functional of $p$ and $f$ is linear of $\Theta$. While sampling methods could lead to more accurate results, variational methods are usually more efficient and they also have an objective to monitor the convergence behavior. Here, we introduce a simple variational method to explore such properties, which turns out to perform well in practice. Specifically, we make the truncated mean-field assumption

$$p(\boldsymbol{\nu}, Z, \Theta) = p(\Theta) \prod_{k=1}^{K} p(\nu_k | \gamma_k) \left( \prod_{i=1}^{N} p(Z_{ik} | \psi_{ik}) \right), \quad (9)$$

where $p(\nu_k | \gamma_k) = \mathrm{Beta}(\gamma_{k1}, \gamma_{k2})$, $p(Z_{ik} | \psi_{ik}) = \mathrm{Bernoulli}(\psi_{ik})$ and $K$ is a truncation level. Then,

problem (8) can be solved using an iterative procedure that alternates between:

**Solving for $p(\Theta)$**: by fixing $p(\nu, Z)$, the subproblem can be equivalently written in a constrained form

$$\min_{p(\Theta),\xi} \mathrm{KL}(p(\Theta)\|p_0(\Theta)) + C \sum_{(i,j)\in\mathcal{I}} \xi_{ij} \quad (10)$$

$\forall (i,j) \in \mathcal{I}$, s.t. : $Y_{ij}(\mathrm{Tr}(\mathbb{E}[W]\bar{\mathbf{Z}}_{ij}) + \mathbb{E}[\eta]^\top X_{ij}) \geq \ell - \xi_{ij}$,

where $\bar{\mathbf{Z}}_{ij} = \mathbb{E}_p[Z_j^\top Z_i]$ is the expected latent features and $\boldsymbol{\xi} = \{\xi_{ij}\}$ are slack variables. By Lagrangian duality theory, we have the optimal solution

$$p(\Theta) \propto p_0(\Theta) \exp\Big\{ \sum_{(i,j)\in\mathcal{I}} \omega_{ij} Y_{ij}(\mathrm{Tr}(W\bar{\mathbf{Z}}_{ij}) + \eta^\top X_{ij}) \Big\}.$$

where $\boldsymbol{\omega} = \{\omega_{ij}\}$ are Lagrangian multipliers.

For the commonly used standard normal prior $p_0(\Theta)$, we have the optimal solution

$$p(\Theta) = p(W)p(\eta) = \Big( \prod_{kk'} \mathcal{N}(\Lambda_{kk'}, 1) \Big) \Big( \prod_d \mathcal{N}(\kappa_d, 1) \Big),$$

where the means are $\Lambda_{kk'} = \sum_{(i,j)\in\mathcal{I}} \omega_{ij} Y_{ij} \mathbb{E}[Z_{ik} Z_{jk'}]$ and $\kappa_d = \sum_{(i,j)\in\mathcal{I}} \omega_{ij} Y_{ij} X_{ij}^d$. The dual problem is

$$\max_{\boldsymbol{\omega}} \sum_{(i,j)} \ell \omega_{ij} - \frac{1}{2}(\|\Lambda\|_2^2 + \|\kappa\|_2^2)$$

$$\text{s.t.} : 0 \leq \omega_{ij} \leq C, \ \forall (i,j) \in \mathcal{I}.$$

Equivalently, the mean parameters $\Lambda$ and $\kappa$ can be directly obtained by solving the primal problem

$$\min_{\Lambda,\kappa,\xi} \frac{1}{2}(\|\Lambda\|_2^2 + \|\kappa\|_2^2) + C \sum_{(i,j)\in\mathcal{I}} \xi_{ij} \quad (11)$$

$\forall (i,j) \in \mathcal{I}$, s.t. : $Y_{ij}(\mathrm{Tr}(\Lambda \bar{\mathbf{Z}}_{ij}) + \kappa^\top X_{ij}) \geq \ell - \xi_{ij}$,

which is a binary classification SVM. We can solve it with any existing high-performance solvers, such as SVMLight or LibSVM.

**Solving for $p(\nu, Z)$**: by fixing $p(\Theta)$, the subproblem is

$$\min_{p(\nu,Z)} \mathrm{KL}(p(\nu,Z)\|p_0(\nu,Z)) + C\mathcal{R}_\ell(p(Z,\Theta)).$$

With the truncated mean-field assumption, we have

$$\mathrm{Tr}(\Lambda \bar{\mathbf{Z}}_{ij}) = \begin{cases} \psi_i \Lambda \psi_j^\top & \text{if } i \neq j \\ \psi_i \Lambda \psi_i^\top + \sum_k \Lambda_{kk} \psi_{ik}(1-\psi_{ik}) & \text{if } i = j \end{cases}$$

We defer the evaluation of the KL-divergence to Appendix A. For $p(\nu)$, since the margin constraints are not dependent on $\nu$, we can get the same solutions as in (Doshi-Velez et al., 2009).

We can solve for $p(Z)$ using sub-gradient methods. Let

$$\mathcal{I}_i = \{j : j \neq i, \ (i,j) \in \mathcal{I} \text{ and } Y_{ij} f(X_{ij}) \leq \ell\}$$
$$\mathcal{I}'_i = \{j : j \neq i, \ (j,i) \in \mathcal{I} \text{ and } Y_{ji} f(X_{ji}) \leq \ell\}.$$

Due to the fact that $\partial_x h_\ell(g(x))$ equals to $-\partial_x g(x)$ if $g(x) \leq \ell$; 0 otherwise, we have the subgradient

$$\partial_{\psi_{ik}} \mathcal{R}_\ell = - \sum_{j \in \mathcal{I}_i} Y_{ij} \Lambda_{k\cdot} \psi_j^\top - \sum_{j \in \mathcal{I}'_i} Y_{ji} \psi_j \Lambda_{\cdot k}$$
$$- \mathbb{I}(Y_{ii} f(X_{ii}) \leq \ell) Y_{ii}(\Lambda_{kk}(1 - \psi_{ik}) + \Lambda_{k\cdot} \psi_i^\top),$$

where $\Lambda_{k\cdot}$ ($\Lambda_{\cdot k}$) denotes the $k$th row (column) of $\Lambda$, and $\mathbb{I}(\cdot)$ is an indicator function. Let the subgradient equal to 0, and we get the update equation

$$\psi_{ik} = \Phi\Big( \sum_{j=1}^k \mathbb{E}_p[\log \nu_j] - \mathcal{L}_k^\nu + C\partial_{\psi_{ik}} \mathcal{R}_\ell \Big). \quad (12)$$

where $\mathcal{L}_k^\nu$ is a lower bound of $\mathbb{E}_p[\log(1 - \prod_{j=1}^k \nu_j)]$ (See Appendix A).

### 3.3. The Fully-Bayesian Model

MedLFRM has one regularization parameter $C$, which normally plays an important role in large-margin classifiers, especially on sparse and imbalanced datasets. To search a good value of $C$, cross-validation is a typical approach, but it could be computationally expensive by comparing many candidates. Under the probabilistic formulation, we could provide an alternative way to control model complexity automatically, at least in part. Below, we present a fully-Bayesian MedLFRM model by introducing appropriate priors for the hyper-parameters.

**Normal-Gamma Prior**: For simplicity, we assume that the prior is an isotropic normal distribution[1] with common mean $\mu$ and precision $\tau$

$$p_0(\Theta|\mu,\tau) = \prod_{kk'} \mathcal{N}(\mu, \tau^{-1}) \prod_d \mathcal{N}(\mu, \tau^{-1}). \quad (13)$$

To complete the model, we use a Normal-Gamma hyper-prior for $\mu$ and $\tau$:

$$p_0(\mu|\tau) = \mathcal{N}(\mu_0, (n_0\tau)^{-1}), \ p_0(\tau) = \mathcal{G}(\frac{\nu_0}{2}, \frac{2}{S_0}), \quad (14)$$

where $\mathcal{G}$ is the Gamma distribution, $\mu_0$ is the prior mean, $\nu_0$ is the prior degrees of freedom, $n_0$ is the prior sample size, $S_0$ is the prior sum of squared errors. We denote this Normal-Gamma distribution by $\mathcal{NG}(\mu_0, n_0, \frac{\nu_0}{2}, \frac{2}{S_0})$.

We note that the normal-Gamma prior has been used in a marginalized form as a heavy-tailed prior for deriving sparse estimates (Griffin & Brown, 2010). Here, we use it for automatically inferring the regularization constants, which replace the role of $C$ in problem (8). Also, our Bayesian approach is different from Max-Margin Nonparametric Latent Feature Models for Link Predictionthe previous methods that were developed for estimating the hyper-parameters of SVM, by optimizing a log-evidence (Gold et al., 2005) or an estimate of the generalization error (Chapelle et al., 2002).

---

[1] A more flexible prior will be the one that uses different means and variances for different components of $\Theta$. We leave this extension for future work.

Formally, with the above hierarchical prior, we define Bayesian MedLFRM (BayesMedLFRM) as solving

$$\min_{p(\boldsymbol{\nu},Z,\mu,\tau,\Theta)} \left\{ \begin{array}{c} \mathrm{KL}(p(\boldsymbol{\nu},Z,\mu,\tau,\Theta) \| p_0(\boldsymbol{\nu},Z,\mu,\tau,\Theta)) \\ +\mathcal{R}_\ell(p(Z,\Theta)) \end{array} \right\}$$

where $p_0(\boldsymbol{\nu},Z,\mu,\tau,\Theta) = p_0(\boldsymbol{\nu},Z)p_0(\mu,\tau)p_0(\Theta|\mu,\tau)$. For this problem, we can develop a similar iterative algorithm as for MedLFRM. Specifically, the sub-step of inferring $p(\boldsymbol{\nu},Z)$ does not change. For $p(\mu,\tau,\Theta)$, the sub-problem (in equivalent constrained form) is

$$\min_{p(\mu,\tau,\Theta),\boldsymbol{\xi}} \mathrm{KL}(p(\mu,\tau,\Theta) \| p_0(\mu,\tau,\Theta)) + \sum_{(i,j)\in\mathcal{I}} \xi_{ij}$$

$$\forall (i,j) \in \mathcal{I}, \text{s.t.} : Y_{ij}(\mathrm{Tr}(\mathbb{E}[W]\bar{\mathbf{Z}}_{ij}) + \mathbb{E}[\eta]^\top X_{ij}) \geq \ell - \xi_{ij},$$

which is convex but intractable to solve directly. Here, we make the mild mean-field assumption that $p(\mu,\tau,\Theta) = p(\mu,\tau)p(\Theta)$. Then, we iteratively solve for $p(\Theta)$ and $p(\mu,\tau)$, as summarized below. We defer the details to Appendix B.

For $p(\Theta)$, we have the mean-field update equation

$$p(W_{kk'}) = \mathcal{N}(\Lambda_{kk'}, \lambda^{-1}), \ p(\eta_d) = \mathcal{N}(\kappa_d, \lambda^{-1}), \quad (15)$$

where $\Lambda_{kk'} = \mathbb{E}[\mu] + \lambda^{-1} \sum_{(i,j)\in\mathcal{I}} \omega_{ij} Y_{ij} \mathbb{E}[Z_{ik}Z_{jk'}]$, $\kappa_d = \mathbb{E}[\mu] + \lambda^{-1} \sum_{(i,j)\in\mathcal{I}} \omega_{ij} Y_{ij} X_{ij}^d$, and $\lambda = \mathbb{E}[\tau]$. Similar as in MedLFRM, the mean of $\Theta$ can be obtained by solving the following problem

$$\min_{\Lambda,\kappa,\boldsymbol{\xi}} \frac{\lambda}{2}(\|\Lambda - \mathbb{E}[\mu]E\|_2^2 + \|\kappa - \mathbb{E}[\mu]\mathbf{e}\|_2^2) + \sum_{(i,j)\in\mathcal{I}} \xi_{ij}$$

$$\text{s.t.} : Y_{ij}(\mathrm{Tr}(\Lambda\bar{\mathbf{Z}}_{ij}) + \kappa^\top X_{ij}) \geq \ell - \xi_{ij}, \ \forall(i,j) \in \mathcal{I},$$

where $\mathbf{e}$ is a $K \times 1$ vector with all entries being the unit 1 and $E = \mathbf{e}\mathbf{e}^\top$ is a $K \times K$ matrix. Let $\Lambda' = \Lambda - \mathbb{E}[\mu]E$ and $\kappa' = \kappa - \mathbb{E}[\mu]\mathbf{e}$, we have the transformed problem

$$\min_{\Lambda',\kappa',\boldsymbol{\xi}} \frac{\lambda}{2}(\|\Lambda'\|_2^2 + \|\kappa'\|_2^2) + \sum_{(i,j)\in\mathcal{I}} \xi_{ij} \quad (16)$$

$$\forall (i,j) \in \mathcal{I}, \text{s.t.} : Y_{ij}(\mathrm{Tr}(\Lambda'\bar{\mathbf{Z}}_{ij}) + (\kappa')^\top X_{ij}) \geq \ell_{ij} - \xi_{ij}$$

where $\ell_{ij} = \ell - \mathbb{E}[\mu]Y_{ij}(\mathrm{Tr}(E\bar{\mathbf{Z}}_{ij}) + \mathbf{e}^\top X_{ij})$ is the adaptive cost. The problem can be solved using an existing binary SVM solver with slight changes to consider the sample-varying costs. Comparing with problem (11), we can see that BayesMedLFRM automatically infers the regularization constant $\lambda$ (or equivalently $C$), by iteratively updating the posterior distribution $p(\tau)$, as explained below.

The mean-field update equation for $p(\mu,\tau)$ is

$$p(\mu,\tau) = \mathcal{N}\mathcal{G}(\tilde{\mu}, \tilde{n}, \tilde{\nu}, \tilde{S}), \quad (17)$$

where $\tilde{\mu} = \frac{K^2\bar{\Lambda} + D\bar{\kappa} + n_0\mu_0}{K^2 + D + n_0}$, $\tilde{n} = n_0 + K^2 + D$, $\tilde{\nu} = \nu_0 + K^2 + D$,

$$\tilde{S} = \mathbb{E}[S_W] + \mathbb{E}[S_\eta] + S_0 + \frac{n_0(K^2(\bar{\Lambda} - \mu)^2 + D(\bar{\kappa} - \mu)^2)}{K^2 + D + n_0},$$

and $S_W = \|W - \bar{W}E\|_2^2$, $S_\eta = \|\eta - \bar{\eta}\mathbf{e}\|_2^2$. From $p(\mu,\tau)$, we can compute the expectation and variance, which are needed in updating $p(\Theta)$

$$\mathbb{E}[\mu] = \tilde{\mu}, \ \mathbb{E}[\tau] = \frac{\tilde{\nu}}{\tilde{S}}, \text{ and } \mathrm{Var}(\mu) = \frac{\tilde{S}}{\tilde{n}(\tilde{\nu} - 2)}.$$

## 4. Experiments

Now, we provide empirical studies on several real datasets to demonstrate the effectiveness of the max-margin principle in learning latent feature relational models, as well as the effectiveness of fully-Bayesian methods in avoiding tuning the hyper-parameter $C$.

### 4.1. Multi-relational Datasets

We report the results of MedLFRM and BayesMedLFRM on the two benchmark datasets which were used in (Miller et al., 2009) to evaluate the performance of latent feature relational models. One dataset contains 54 relations of 14 countries along with 90 given features of the countries, and the other one contains 26 kinship relationships of 104 people in the Alyawarra tribe in Central Australia. On average, there is a probability of about 0.21 that a link exists for each relation on the countries dataset, and the probability of a link is about 0.04 for the kinship dataset. So, the kinship dataset is extremely imbalanced (i.e., much more negative examples than positive examples). To deal with this imbalance in learning max-margin MedLFRM, we use different regularization constants for the positive ($C^+$) and negative ($C^-$) examples. We refer the readers to (Akbani et al., 2004) for other possible choices. In our experiments, we set $C^+ = 10C^- = 10C$ for simplicity and tune the parameter $C$. For BayesMedLFRM, this equality is held during all iterations.

Depending on the input data, the latent features might not have interpretable meanings (Miller et al., 2009). In the experiments, we focus on the effectiveness of max-margin learning in learning latent feature relational models. We also compare with two well-established class-based algorithms – IRM (i.e., infinite relational model) (Kemp et al., 2006) and MMSB (i.e., mixed membership stochastic block) (Airoldi et al., 2008), both of which were tested in (Miller et al., 2009). In order to compare with their reported results, we use the same setup for the experiments. Specifically, for each dataset, we held out 20% of the data during training and report the AUC (i.e., area under the Receiver Operating Characteristic or ROC curve) for the held out data. As in (Miller et al., 2009), we consider two settings – "global" and "single". For the global setting, we infer a single set of latent features for all relations; and for the single setting, we infer independent latent features for each relation and the overall AUC is an average of the AUC scores of all relations.

Table 1. AUC on the countries and kinship datasets. Bold indicates the best performance.

| | Countries single | Countries global | Alyawarra single | Alyawarra global |
|---|---|---|---|---|
| SVM | 0.8180 ± 0.0000 | 0.8180 ± 0.0000 | — | — |
| LR | 0.8139 ± 0.0000 | 0.8139 ± 0.0000 | — | — |
| MMSB | 0.8212 ± 0.0032 | 0.8643 ± 0.0077 | 0.9005 ± 0.0022 | 0.9143 ± 0.0097 |
| IRM | 0.8423 ± 0.0034 | 0.8500 ± 0.0033 | 0.9310 ± 0.0023 | 0.8943 ± 0.3000 |
| LFRM rand | 0.8529 ± 0.0037 | 0.7067 ± 0.0534 | 0.9443 ± 0.0018 | 0.7127 ± 0.0300 |
| LFRM w/ IRM | 0.8521 ± 0.0035 | 0.8772 ± 0.0075 | 0.9346 ± 0.0013 | 0.9183 ± 0.0108 |
| MedLFRM | **0.9173** ± 0.0067 | **0.9255** ± 0.0076 | **0.9552** ± 0.0065 | **0.9616** ± 0.0045 |
| BayesMedLFRM | **0.9178** ± 0.0045 | **0.9260** ± 0.0023 | **0.9547** ± 0.0028 | **0.9600** ± 0.0016 |

For MedLFRM and BayesMedLFRM, we randomly initialize the posterior mean of $W$ uniformly in the interval $[0, 0.1]$; initialize $\psi$ to uniform (i.e., 0.5) corrupted by a random noise distributed uniformly at the interval $[0, 0.001]$; and initialize the mean of $\eta$ to be zero. All the following results of MedLFRM and BayesMedLFRM are averages over 5 randomly initialized runs, again similar as in (Miller et al., 2009). For MedLFRM, the hyper-parameter $C$ is selected via cross-validation during training. For BayesMedLFRM, we use a very weak hyper-prior by setting $\mu_0 = 0$, $n_0 = 1$, $\nu_0 = 2$, and $S_0 = 1$. We set the cost parameter $\ell = 9$ in all experiments.

Table 1 shows the results. We can see that in both settings and on both datasets, the max-margin based latent feature relational model MedLFRM significantly outperforms LFRM that uses a likelihood-based approach with MCMC sampling. Comparing BayesMedLFRM and MedLFRM, we can see that using the fully-Bayesian technique with a simple Normal-Gamma hierarchical prior, we can avoid tuning the regularization constant $C$, without sacrificing the link prediction performance. To see the effectiveness of latent feature models, we also report the performance of logistic regression (LR) and linear SVM on the countries dataset, which has input features. We can see that a latent feature or latent class model generally outperforms the methods that are built on raw input features for this particular dataset.

Figure 1 shows the performance of MedLFRM on the countries dataset when using and not using input fea-tures. We consider the global setting. Here, we also study the effects of truncation level $K$. We can see that in general using input features can boost the performance. Moreover, even if using latent features only, MedLFRM can still achieve very competitive performance, better than the performance of the likelihood-based LFRM that uses both latent features and input features. Finally, it is sufficient to get good performance by setting the truncation level $K$ to be larger than 40. We set $K$ to be 50 in the experiments.

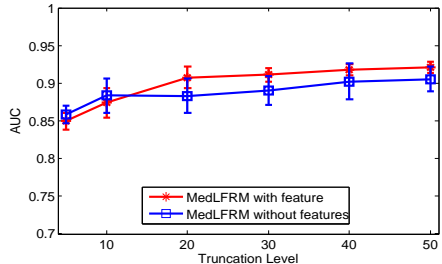

Figure 1. AUC scores of MedLFRM with and without input features on the countries dataset.

### 4.2. Predicting NIPS coauthorship

The second experiments are done on the coauthorship data constructed from the NIPS dataset which contains a list of all papers and authors from NIPS 1-17. To compare with LFRM, we use the same dataset as in (Miller et al., 2009), which contains 234 authors who had published with the most other people[2]. To better fit the symmetric coauthor link data, we restrict our models to be symmetric, i.e., the posterior mean of $W$ is a symmetric matrix, as in (Miller et al., 2009). For MedLFRM and BayesMedLFRM, this symmetry constraint can be easily satisfied when solving the SVM problems (11) and (16). To see the effects of the symmetry constraint, we also report the results of the asymmetric MedLFRM and asymmetric BayesMedLFRM, which do not impose the symmetry constraint on the posterior mean of $W$. As in (Miller et al., 2009), we train the model on 80% of the data and use the remaining data for test.

---

[2]The average probability of forming a link on this data is about 0.02, again very imbalanced. We tried the same strategy as for the kinship dataset by using different regularization constants. The results are not significantly different from those by using a common $C$. $K = 80$ is sufficient for these experiments.

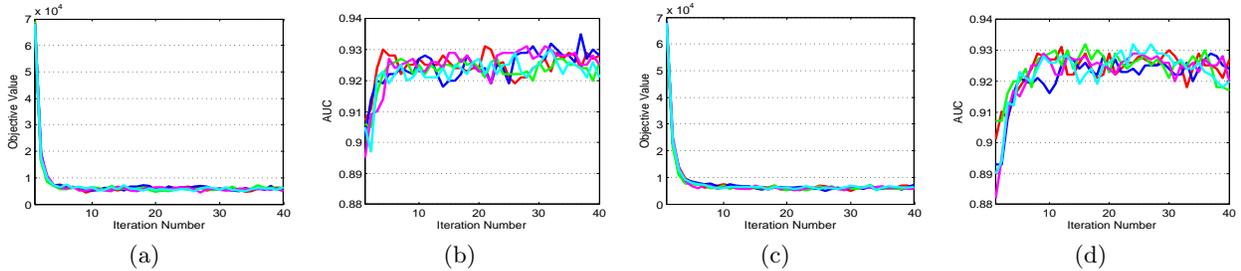

Figure 2. (a-b) Objective values and test AUC during iterations for MedLRFM; and (c-d) objective values and test AUC during iterations for Bayesian MedLRFM on the countries dataset with 5 randomly initialized runs.

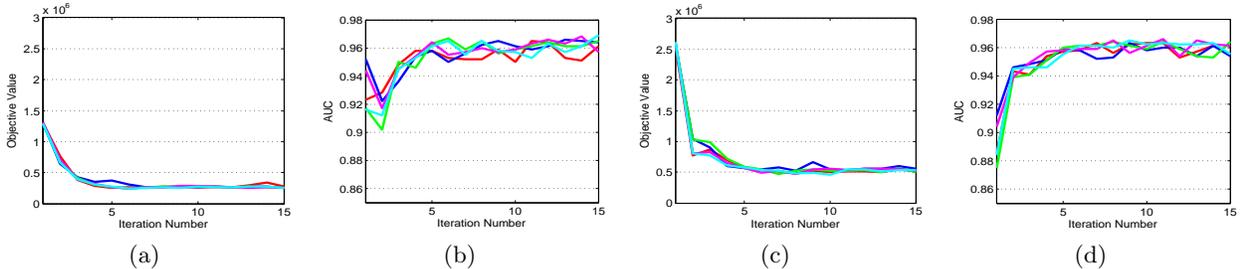

Figure 3. (a-b) Objective values and test AUC during iterations for MedLRFM; and (c-d) objective values and test AUC during iterations for Bayesian MedLRFM on the kinship dataset with 5 randomly initialized runs.

Table 2. AUC on the NIPS coauthorship data. Bold indicates the best performance.

| | |
|---|---|
| MMSB | $0.8705 \pm 0.0130$ |
| IRM | $0.8906 \pm -$ |
| LFRM rand | $0.9466 \pm -$ |
| LFRM w/ IRM | $0.9509 \pm -$ |
| MedLFRM | **$0.9642 \pm 0.0026$** |
| BayesMedLFRM | **$0.9636 \pm 0.0036$** |
| Asymmetric MedLFRM | $0.9140 \pm 0.0130$ |
| Asymmetric BayesMedLFRM | $0.9146 \pm 0.0047$ |

Table 2 shows the results, where the results of LFRM, IRM and MMSB were reported in (Miller et al., 2009). Again, we can see that using the discriminative max-margin training, the symmetric MedLFRM and BayesMedLFRM outperform all other likelihood-based methods, using either latent feature or latent class models; and the fully-Bayesian MedLFRM model performs comparably with MedLFRM while avoiding tuning the hyper-parameter $C$. Finally, the asymmetric MedLFRM and BayesMedLFRM models perform much worse than their symmetric counterpart models, but still better than the latent class models.

### 4.3. Stability and Running Time

Figure 2 shows the change of the objective function as well as the change of the AUC scores on test data of the countries dataset during the iterations for both MedLFRM and BayesMedLFRM. For MedLFRM, we report the results with the best $C$ selected via cross-validation. We can see that the variational inference algorithms for both models converge quickly to a particular region. Since we use sub-gradient descent to update the distribution of $Z$ and the subproblems of solving for $p(\Theta)$ can in practice only be approximately solved, the objective function has some disturbance, but within a relatively very small interval. For the AUC scores, we have similar observations, namely, within several iterations, we could have very good link prediction performance. The disturbance is again maintained within a small region, which is reasonable for our approximate inference algorithms. Comparing the two models, we can see that BayesMedLFRM has similar behaviors as MedLFRM, which demonstrates the effectiveness of using fully-Bayesian techniques to automatically learn the hyper-parameter $C$. Figure 3 presents the results on the kinship dataset, from which we have the same observations. We omit the results on the NIPS dataset for saving space.

Finally, Figure 4 shows the training time and test time of MedLFRM and Bayesian MedLFRM on each of the three datasets. For MedLFRM, we show the single run with the optimum parameter $C$, selected via inner cross-validation. We can see that using Bayesian infer-ence, the running time does not increase much, being generally comparable with that of MedLFRM. But since MedLFRM needs to select the hyper-parameter $C$, it will need much more time than BayesMedLFRM to finish the entire training on a single dataset.

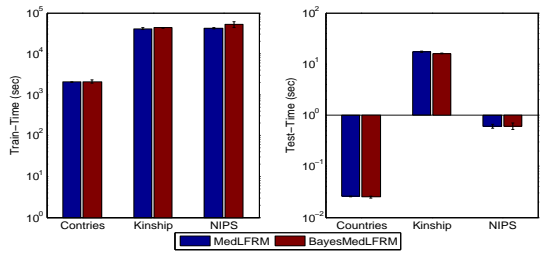

*Figure 4.* Training and test time on different datasets.

## 5. Conclusions and Future Work

We have presented a discriminative max-margin latent feature relational model for link prediction. Under a Bayesian-style max-margin formulation, our work naturally integrates the ideas of Bayesian nonparametrics to automatically resolve the unknown dimensionality of a latent social space. Furthermore, we present a fully-Bayesian formulation, which can avoid tuning regularization constants. We developed efficient variational methods to perform posterior inference. Empirical results on several real datasets appear to demonstrate the benefits inherited from both max-margin learning and fully-Bayesian methods.

Our current analysis is focusing on small static network snapshots. For future work, we are interested in learning more flexible latent feature relational models to deal with large dynamic networks and reveal more subtle network evolution patterns. We are also interested in developing Monte Carlo sampling methods, which have been widely used in previous latent feature relational models.

## Acknowledgements

JZ is supported by National Key Foundation R&D Projects 2012CB316301, a Starting Research Fund No. 553420003, the 221 Basic Research Plan for Young Faculties at Tsinghua University, and a Research Fund No. 20123000007 from Microsoft Research Asia.